\documentclass[10pt]{article}
\pdfoutput=1
\usepackage[T1]{fontenc}
\usepackage{times}
\usepackage{amsmath,amssymb,amsfonts}
\usepackage{amsthm}
\usepackage{booktabs}
\usepackage{multirow}
\usepackage{graphicx}
\usepackage{microtype}
\usepackage{enumitem}
\usepackage{mathtools}
\usepackage{tabularx}
\usepackage{threeparttable}
\usepackage{url}
\usepackage{seqsplit}
\usepackage{algorithm}
\usepackage{algpseudocode}
\usepackage{tikz}
\usetikzlibrary{arrows.meta,positioning,fit,calc}
\usepackage[preprint]{tmlr}
\usepackage[capitalize,noabbrev,nameinlink]{cleveref}

\theoremstyle{plain}
\newtheorem{proposition}{Proposition}[section]

\newtheorem{assumption}[proposition]{Assumption}
\theoremstyle{definition}
\newtheorem{definition}[proposition]{Definition}
\newtheorem{remark}[proposition]{Remark}

\algrenewcommand\algorithmicrequire{\textbf{Input:}}
\algrenewcommand\algorithmicensure{\textbf{Output:}}
\algrenewcommand\algorithmiccomment[1]{\hfill$\triangleright$~#1}

\newcommand{\D}{\mathcal{D}}
\newcommand{\F}{\mathcal{F}}

\newcommand{\Pplan}{\mathcal{P}}
\newcommand{\TState}{\mathsf{S}}
\newcommand{\TraceDelete}{\mathsf{TD}}
\newcommand{\Repack}{\mathsf{RP}}

\newcommand{\hashstr}[1]{{\small\texttt{\seqsplit{#1}}}}

\title{Unlearning at Scale: State-Exact Trace-Preserving Deletion in Billion-Parameter Language Models}

\author{\name Abdullah X \\
\email Abdullah@projectawareAI.org \\
\addr Project AWARE}

\begin{document}
\maketitle

\begin{abstract}
Can a prospectively instrumented training continuation reproduce a deletion counterfactual exactly after selected examples leave its replay dataset? We study a trace-preserving counterfactual that fixes the recorded execution controls while assigning requested identifiers zero contribution. The guarantee is prospective: the original run must record this execution provenance and retain an eligible uncontaminated checkpoint. Under pinned single-GPU environments, replay from a token store materialized without the requested rows reconstructs a separately executed trace oracle bit-for-bit in model and optimizer state. Pythia 160M is exact across four deletion geometries; Pythia 2.8B matches all 2,775,208,960 model-state elements for a random 5\% request; and Llama 3.2 1B is exact after omitting 400 of 4,000 TOFU examples from replay storage. These results establish billion-parameter state exactness. They do not establish cheap deletion, because dispersed requests can force nearly full replay. We release standardized TOFU/OpenUnlearning measurements as descriptive diagnostics only because the frozen campaign lacks the matched controls required for a causal behavioral claim.
\end{abstract}

\section{Introduction}

Requests to erase data after model training create an unusually concrete systems question: what computation should replace a model after a deletion request targets records that influenced its parameters? Article 17 of the European Union's General Data Protection Regulation provides an important legal motivation for erasure requests, although the legal meaning and sufficiency of any technical operation depend on context and are outside the scope of this paper~\citep{gdpr2016}. Language models sharpen this problem because they can memorize and sometimes reproduce training data~\citep{carlini2021,carlini2022memorization}. Contemporary training pipelines also behave as stateful, stochastic, hardware-dependent programs, so an unordered dataset alone does not determine the realized training state.

Machine-unlearning research has long used retraining on the retained data as a reference point~\citep{cao2015,ginart2019,guo2020}. This reference is conceptually powerful, but a concrete deterministic state-equality claim requires more than naming the retained examples. A run of AdamW over $\D\setminus\F$ depends on their order and grouping, random-number consumption, accumulation boundaries, learning-rate schedule, optimizer state, numerical kernels, and floating-point reduction order. If deletion causes retained examples to be densely repacked, the resulting training program differs from one that preserves the original logical positions. Formal definitions can avoid this ambiguity by treating the learner as a fully specified randomized algorithm or output distribution; our concern is the realized state of an implemented training execution. For that state-level question, ``the model trained without $\F$'' does not identify a unique reference state unless the realized execution or reference distribution is specified~\citep{block2025}.

This paper makes that specification explicit. We define a \emph{trace-preserving deletion counterfactual}: the original training execution fixes logical microbatch slots, per-microbatch random seeds, accumulation boundaries, optimizer-step indices, and learning-rate values. A delete request sets the loss contribution of forgotten slots to zero while preserving these control inputs. Crucially, preserving a slot does not require preserving the forgotten content. Before replay, we materialize a new token store that omits the requested rows. When a forgotten identifier appears in the immutable execution plan, replay substitutes a fixed dummy sequence with zero sample weight before attempting any row lookup. The requested rows are therefore absent from the replay token store and contribute no training loss. This is \emph{prospective unlearning infrastructure}: the original continuation must have been run under the replay-aware contract and with compatible checkpoints. The mechanism does not retrospectively reconstruct an unrecorded execution trace for an arbitrary existing model.

This target is intentionally narrower than universal retraining equivalence. It is \emph{exact to a declared counterfactual program}. That distinction follows the argument of \citet{thudi2022} that auditable unlearning claims must be stated at the algorithmic level, and it complements recent work questioning whether dataset removal, knowledge suppression, refusal behavior, and other objectives should share one label~\citep{cooper2025position,yoon2026overused}. We do not argue that trace preservation is the uniquely correct deletion semantics. We contribute a systems construction for one explicit semantics and ask whether a real training system can maintain its invariants through replay-store redaction at billion-parameter state scale.

The state-level evidence is binary. We do not infer exactness from behavioral similarity. On Pythia 160M, no-deletion replay reproduces all 162,322,944 model elements and the optimizer state exactly. Across early 0.1\%, middle 1\%, late 1\%, and random 5\% requests, replay from a physically redacted token store exactly matches a separately executed trace oracle. Physical filtering and dense repacking produce nonzero state differences in all four cases; we use these only as non-equivalence witnesses, not as evidence that trace preservation is a better unlearning objective. A parameter-scale check on Pythia 2.8B reproduces all 2,775,208,960 model elements and optimizer state exactly for identity replay and a random 5\% deletion. A cross-family experiment on Llama 3.2 1B Instruct and TOFU omits 400 of 4,000 dataset examples from replay storage and exactly reproduces the trace oracle over 1,498,482,688 model elements and optimizer state.

We also reconstruct the frozen Llama states by hash and run the pinned OpenUnlearning TOFU evaluator~\citep{dorna2025openunlearning}. We deliberately treat this as a \emph{secondary descriptive diagnostic}. The official \texttt{retain90} checkpoint comes from a different training trajectory. The frozen campaign also lacks a matched local retained-only control and pre-fine-tuning TOFU behavioral metrics. Consequently, these measurements are not used to infer that trace preservation causes a particular forgetting behavior. Their role is narrower: they document how the released exact states score under a standardized evaluator and reinforce the need to keep state exactness and behavioral objectives conceptually separate.

Our contributions are:
\begin{itemize}[leftmargin=1.4em,itemsep=0.35em,topsep=0.35em]
\item \textbf{An execution-conditioned definition of exact deletion.} We formalize trace-preserving deletion separately from densely repacked retained-data training, making the counterfactual program part of the unlearning specification.
\item \textbf{A prospectively instrumented replay system with replay-store redaction.} The system records a 32-byte fixed-width microbatch WAL plus an ordered-ID manifest, reconstructs deterministic control inputs, and replays after requested token rows have been omitted from the replay token store.
\item \textbf{Bit-exact billion-parameter evidence.} We demonstrate exact model-and-optimizer identity and deletion replay on Pythia 160M, Pythia 2.8B, and Llama 3.2 1B Instruct under pinned single-GPU environments, including a 2.8B-parameter state-scale endpoint and a TOFU cross-family endpoint.
\item \textbf{A replay-locality analysis.} Measured checkpoint choices and an order-statistic calculation show a structural limitation of suffix replay: localized late requests can avoid substantial computation, whereas large randomly dispersed requests are expected to force nearly full replay.
\item \textbf{A frozen, auditable research artifact.} Every publication-facing run records source revisions, prepared-data hashes, execution-plan and WAL hashes, model and optimizer hashes, environment snapshots, workflow identifiers, and compact evidence artifacts in a public release~\citep{abdullah2026artifact}. The release also includes standardized OpenUnlearning measurements, which we use descriptively and without causal attribution.
\end{itemize}

\section{Related Work}
\label{sec:related}

\subsection{Deletion-efficient, exact, and certified unlearning}

Early work framed unlearning as efficiently updating a learned system after deletion requests. \citet{cao2015} proposed systems that support removal by construction, while \citet{ginart2019} formalized deletion-efficient learning. \citet{guo2020} introduced certified removal, requiring an output distribution that is difficult to distinguish from a model that never observed the removed data. \citet{neel2021} sharpened an important distinction between indistinguishability of the learner's complete internal state and indistinguishability only of its observable output. \citet{sekhari2021} and \citet{gupta2021adaptive} developed additional algorithmic guarantees, including adaptive deletion settings. Recent certified approaches extend formal guarantees to neural networks under different mechanisms and assumptions~\citep{koloskova2025}.

SISA training reduces deletion cost by sharding, isolating, slicing, and aggregating models so that only affected components need retraining~\citep{bourtoule2021}. More recent exact-unlearning systems similarly modify the learning architecture to localize influence. S3T isolates data across sequential parameter-efficient components~\citep{chowdhury2025s3t}; LMEraser partitions private data across independently tunable prompts~\citep{xu2025lmeraser}; and \citet{muresanu2025} replace SGD-based fine-tuning with an in-context adaptation scheme that permits efficient exact deletion. These approaches demonstrate that exactness can be made practical by redesigning how adaptation is performed.

Under conventional exact-unlearning definitions, the reference is often an independently rerun learner on retained data or an output distribution close to it~\citep{guo2020,neel2021}. Our trace target makes a different claim. It embeds a conventional gradient-based optimizer and transformer training loop in a replay-aware execution contract with per-microbatch reseeding, fixed additive loss semantics, an immutable plan, and explicit provenance. The system verifies bitwise equality of model and optimizer state against a separately executed trace-preserving oracle. This is closer in spirit to system-aware formulations that make the learner's stored state and threat model explicit~\citep{lu2025system} and to the algorithmic auditability argument of \citet{thudi2022} than to a claim that parameters alone prove historical data absence.

\subsection{LLM unlearning as behavioral optimization}

The LLM literature has largely developed around post-hoc changes to model behavior. Early work applied gradient ascent to target sequences to reduce memorized knowledge~\citep{jang2023knowledge}; broader studies subsequently compared multiple update-based unlearning objectives on pretrained LMs~\citep{yao2024pretrained}. Other formulations move the intervention outside parameter updates. In-context unlearning, for example, changes inference context while leaving model weights fixed~\citep{pawelczyk2024incontext}. A recent review of LLM unlearning emphasizes that the field spans distinct scopes, data--model interactions, objectives, and efficacy criteria~\citep{liu2025rethinking}. That breadth makes the object called ``unlearning'' part of the scientific specification.

TOFU introduced a fictitious-author benchmark with structured forget and retain splits and a suite of behavioral metrics~\citep{maini2024tofu}. Gradient-ascent objectives can destabilize models; Negative Preference Optimization (NPO) was proposed to improve the forgetting--utility trade-off~\citep{zhang2024npo}, and SimNPO removes NPO's dependence on a reference model~\citep{fan2024simnpo}. \citet{yuan2025closer} analyze failure modes and evaluation choices for LLM unlearning. MUSE broadens evaluation to verbatim and knowledge memorization, privacy leakage, retained utility, scale, and sequential deletion~\citep{shi2025muse}. OpenUnlearning then unifies a large set of methods, metrics, and benchmark configurations and performs meta-evaluation of unlearning metrics~\citep{dorna2025openunlearning}.

The methodological space has continued to widen. Accepted 2026 work includes reinforcement-learning formulations of behavioral suppression~\citep{zaradoukas2026purge}, model-editing formulations of targeted knowledge remapping~\citep{lin2026zerounlearn}, and inference-time steering with auxiliary models~\citep{merchant2026divergence}. Complementary evidence shows that how knowledge is encoded during learning can materially change later unlearning behavior~\citep{wu2026encoding}. These methods answer useful questions under different state-level counterfactuals or behavioral objectives. We therefore separate exact-state verification from behavioral evaluation. The pinned OpenUnlearning run is reported only as a reproducible diagnostic of the released states. We omit publication-authoritative comparisons with GradAscent, NPO, SimNPO, and other approximate methods because the release contains no matched runs from the same project target. These claim types require different evidence.

\subsection{Behavioral forgetting and deletion}

A growing evaluation literature warns that successful-looking behavioral forgetting may not imply removal of training influence. \citet{du2025false} show that textual unlearning can leave substantial privacy leakage. \citet{xu2025reversibility} demonstrate that apparently forgotten LLM behavior can sometimes be rapidly restored and propose representation-level analyses of reversibility. The NeurIPS position paper by \citet{cooper2025position} cautions against treating machine unlearning as a general-purpose solution for generative-AI policy goals. Most recently, \citet{yoon2026overused} argue for reserving ``machine unlearning'' for dataset-defined deletion and separating it from suppression, editing, and refusal objectives.

Our evidence approaches this distinction from the state side. We start from a state-exact, replay-store-redacted execution-conditioned deletion and then report standardized behavioral measurements without treating them as an identified effect of trace deletion. The general lesson concerns scope: state exactness and behavioral objectives answer different questions, and causal links between them require matched controls.

\section{Problem Formulation}
\label{sec:formulation}

\subsection{Training is an executed program}

Let $\D=((i,z_i))_{i=1}^n$ be an indexed tokenized record store and let $\F$ be a set of requested identifiers. This distinguishes a dataset example from its repeated presentations in an execution plan. We write the replayed model-and-optimizer state as
\begin{equation}
\TState_t=(\theta_t,\Omega_t),
\end{equation}
where $\theta_t$ is the model state and $\Omega_t$ is optimizer state. This pair represents the model-and-optimizer state under our execution contract. In our contract, RNG choices, learning-rate values, accumulation phase, sample order, and other transition controls are externalized into the plan $\Pplan$ and pinned environment $E$. Equal parameters with different Adam moments still need not define the same future learner, so $\Omega_t$ remains part of the exactness target.

We represent the training execution by an ordered plan
\begin{equation}
\Pplan=(p_1,\ldots,p_M), \qquad
p_j=(I_j,s_j,\eta_j,u_j,a_j),
\label{eq:plan}
\end{equation}
where $I_j$ is the ordered tuple of identifiers occupying microbatch plan record $j$, $s_j$ is its random seed, $\eta_j$ is the learning-rate value, $u_j$ is the logical optimizer step to which the microbatch contributes, and $a_j\in\{0,1\}$ denotes the end of a gradient-accumulation segment. A checkpoint labeled $k$ denotes the state immediately before logical optimizer step $k$, equivalently after completion of step $k-1$. We condition exactness on a pinned implementation environment $E$ containing model/data revisions, software stack, device, deterministic-kernel configuration, and numerical dtype.

The released system uses an \emph{unlearning-aware training contract}. It reseeds Python, NumPy, PyTorch, and CUDA per microbatch, writes the recorded learning rate into the optimizer, and accumulates a sum-reduced weighted token loss. Sum reduction forms part of the counterfactual contract. With mean reduction, deleting an example can change the normalization denominator and therefore rescale retained gradients. Our additive semantics make a forgotten slot remove an addend while leaving retained contributions numerically unchanged. Other normalization rules could define legitimate but different counterfactuals.

\subsection{Trace-preserving deletion}

The counterfactual used by this paper preserves $\Pplan$ but replaces every forgotten logical slot with a dummy sequence having sample weight zero.

\begin{definition}[Trace-preserving deletion]
For a deletion request $\F$, define $\TraceDelete(\Pplan,\F)$ as the execution that retains every microbatch plan record $p_j$ but maps each identifier in $I_j\cap\F$ to a fixed dummy example whose loss weight is zero. Retained identifiers, seeds, learning rates, logical optimizer-step indices, accumulation boundaries, and batch shapes are unchanged. At an accumulation boundary, an optimizer transition is executed if and only if at least one retained weighted contribution occurred in that segment; otherwise the logical boundary is preserved but the optimizer transition is skipped.
\end{definition}

The empty-segment rule is part of the counterfactual. Executing AdamW with zero gradients would be a different semantics because optimizer step counters, moments, bias correction, and decoupled weight decay can change even when no data gradient is present. An \emph{empty-step-preserving} trace is therefore possible, but it would define another target.

Let $\TState_T^{-\F,\mathrm{trace}}$ denote the terminal model-and-optimizer state produced by this counterfactual from an eligible checkpoint. This state is our exact target.

\begin{definition}[Trace exactness]
A deletion procedure is trace-exact for $(\Pplan,\F,E)$ if its terminal state is bit-identical, in the recorded state dtypes, to $\TState_T^{-\F,\mathrm{trace}}$ for both model and optimizer state.
\end{definition}

This definition is stronger than behavioral closeness to the trace oracle but narrower than a universal statement about all possible ways to retrain on $\D\setminus\F$.

\subsection{Physical redaction is a separate property}

The trace can retain an identifier even after the corresponding data row has been deleted. We therefore distinguish logical provenance from data storage.

\begin{definition}[Physical replay redaction]
A replay store is physically redacted for $\F$ if no token row, label row, or metadata record corresponding to an identifier in $\F$ is present in that store. The immutable execution plan may retain those identifiers solely as logical slot markers.
\end{definition}

At replay time the implementation checks the forget-set membership of the identifier before row lookup. Replay synthesizes each forgotten slot locally as a dummy array and assigns it weight zero. Thus the replay algorithm does not need the deleted token content to reconstruct the trace-preserving computation.

\subsection{Dense retained-data retraining is a different target}

A conventional retained-data run can remove all identifiers in $\F$ and densely repack the remaining identifiers into new microbatches. Let $\Repack(\Pplan,\F)$ denote a new plan built from those retained examples. In general,
\begin{equation}
\Repack(\Pplan,\F) \neq \TraceDelete(\Pplan,\F).
\end{equation}
The two programs can differ in batch shapes, RNG consumption, accumulation composition, optimizer-update count, schedule index, and floating-point reduction order. We therefore call the former the \emph{repacked retained-data counterfactual}, not the oracle for our trace-exact claim. The Pythia 160M experiment measures this distinction directly.

\begin{remark}
Nothing in the trace-exact definition proves that a parameter vector reveals whether a record was historically used. That would conflict with the auditability limitation identified by \citet{thudi2022}. Our auditable claim is algorithmic: the released evidence identifies the source state, deletion request, redacted store, execution contract, and equality to the declared counterfactual.
\end{remark}

\section{System Design}
\label{sec:system}

\subsection{Immutable execution plan and write-ahead log}

The original run first builds an immutable JSONL execution plan from the ordered sample identifiers and the training configuration. Per-microbatch seeds are deterministically derived from a base seed and microbatch index using SHA-256; learning-rate values are quantized to float32 when the plan is created. The plan itself is SHA-256 hashed.

During training, the writer records the same information in a compact binary WAL and an ordered-ID manifest. The binary WAL record is exactly 32 bytes (\cref{tab:wal}). A CRC32 detects per-record corruption; SHA-256 records whole-file identities for the WAL and manifest. The manifest stores the full ordered sample identifiers and a full digest, while the fixed-width WAL stores a 64-bit identifier tag. The 64-bit field is a compact consistency tag, not a modern cryptographic identity primitive. The reader verifies record length, CRC, manifest agreement, sample-ID digest, and whole-file hashes before reconstructing the plan.

\begin{table}[t]
\centering
\caption{Fixed-width binary WAL record. Ordered sample identifiers are stored separately in an integrity-checked manifest; the 32-byte value is therefore not the complete provenance footprint.}
\label{tab:wal}
\small
\begin{tabular}{lrl}
\toprule
Field & Bytes & Purpose \\
\midrule
Ordered-ID tag & 8 & Compact consistency tag for ordered IDs \\
Microbatch seed & 8 & Reconstructs stochastic draws \\
Learning rate & 4 & Float32 value applied during replay \\
Logical optimizer step & 4 & Preserves update identity \\
Microbatch length & 2 & Integrity and shape metadata \\
Flags & 2 & Accumulation-end boundary \\
CRC32 & 4 & Per-record corruption detection \\
\midrule
Total & 32 & Fixed binary record only \\
\bottomrule
\end{tabular}
\end{table}

\subsection{Deterministic execution contract}

The released experiments fix the Python, NumPy, PyTorch, and CUDA seeds. They disable cuDNN benchmarking and TF32, request deterministic cuDNN behavior and strict PyTorch deterministic algorithms, set \texttt{CUBLAS\_WORKSPACE\_CONFIG=:4096:8}, and reseed the RNGs at every microbatch. PyTorch itself cautions that complete reproducibility is not guaranteed across arbitrary releases, platforms, or devices~\citep{pytorch2026randomness}. Accordingly, the environment is an explicit part of our exactness contract, and hardware-independent determinism lies outside the claim.

The original and replay paths load the same pinned model revision and checkpoint, use the same optimizer configuration, and set the optimizer learning rate directly from the plan before every contributing microbatch. At an accumulation boundary, the optimizer transition follows the rule in the trace-deletion definition: advance only if the segment contains retained weighted contribution; otherwise preserve the logical boundary and skip the transition. These rules are prospective design choices. We generally cannot give this exact replay guarantee after the fact to a conventionally trained model that lacks the required plan, per-microbatch controls, and eligible checkpoints.

\subsection{Deletion oracle and physically redacted replay}

For each deletion scenario, the runner finds the earliest logical optimizer step touched by the forget set and selects the latest eligible checkpoint whose label is no greater than that step. Because checkpoint $k$ is the state immediately before logical step $k$, such a checkpoint contains no contribution from the first affected step. The \emph{trace oracle} is then executed as a separate run from that checkpoint using the original plan with forgotten slots mapped to zero-weight dummies. Oracle and replay substantially share the same transition implementation; ``separate'' therefore refers to execution, not independent implementation. We freeze the oracle's terminal model and optimizer hashes.

The replay path then materializes a new token store containing only retained rows. The store writes a redaction manifest recording source, forgotten, and retained counts and whether any forgotten identifier remains. Replay loads the same checkpoint and reconstructed plan, but reads retained examples from the physically redacted store. Forgotten logical slots are intercepted before data lookup and replaced with the same dummy representation used by the oracle.

\begin{figure}[t]
\centering
\resizebox{0.98\textwidth}{!}{%
\begin{tikzpicture}[
  node distance=6mm and 6mm,
  box/.style={draw,rounded corners,align=center,inner sep=5pt,font=\small,minimum height=8mm},
  emph/.style={box,very thick},
  arr/.style={-{Latex[length=2mm]},thick}
]
\node[box] (original) {Original training\\$\D,\Pplan,E$};
\node[box,right=of original] (trace) {Frozen plan + WAL\\IDs, seeds, LR, steps, boundaries};
\node[box,right=of trace] (ckpt) {Eligible checkpoint\\$\TState_k$};
\draw[arr] (original) -- (trace);
\draw[arr] (trace) -- (ckpt);
\node[emph,below left=10mm and -3mm of ckpt] (oracle) {Trace-preserving oracle\\forget slots $\rightarrow$ zero weight};
\node[emph,below right=10mm and -3mm of ckpt] (replay) {Physically redacted replay\\deleted rows absent};
\draw[arr] (ckpt) -- (oracle);
\draw[arr] (ckpt) -- (replay);
\node[box,below=9mm of oracle] (os) {Oracle state\\$(\theta_T,\Omega_T)$};
\node[box,below=9mm of replay] (rs) {Replay state\\$(\theta'_T,\Omega'_T)$};
\draw[arr] (oracle) -- (os);
\draw[arr] (replay) -- (rs);
\draw[<->,very thick] (os) -- node[below,font=\small,align=center]{bit-identical\\model + optimizer} (rs);
\node[box,left=12mm of oracle] (repack) {Dense retained-data repack\\new plan $\Repack(\Pplan,\F)$};
\draw[arr,dashed] (ckpt.south west) -- (repack.north east);
\node[font=\footnotesize,align=center,below=2mm of repack] {alternative retained-data\\counterfactual, not expected to match};
\end{tikzpicture}%
}
\caption{The exactness claim conditions on an explicit execution counterfactual. Oracle and redacted replay use the same checkpoint and immutable plan; dense retained-data repacking defines another program.}
\label{fig:system}
\end{figure}

\subsection{The systems invariant}

The equality proposition below is intentionally elementary. If two deterministic executions begin from the same state and receive identical numerical transition inputs, equality follows by induction. The research question is whether a prospectively instrumented system can maintain that invariant after replay-store redaction through real transformer training at billion-parameter state scale. Oracle and replay are separately executed but substantially share the transition code, so the experiment verifies a systems invariant. It does not test an independently implemented oracle.

\begin{assumption}[Pinned deterministic transition]
For a fixed environment $E$, model/optimizer state, numerical microbatch, seed, learning rate, and accumulation boundary, the training transition is deterministic in the recorded state dtypes.
\label{ass:det}
\end{assumption}

\begin{assumption}[Eligible checkpoint]
The starting checkpoint precedes the first contribution of every requested record in the replayed suffix.
\label{ass:checkpoint}
\end{assumption}

\begin{assumption}[Replay semantics]
The oracle and replay use the same immutable plan and the same dummy representation for every forgotten slot. The loss is sum-reduced over valid tokens after multiplication by per-sample weights, with forgotten slots assigned weight zero. At each accumulation boundary both executions apply an optimizer transition if and only if that segment contains at least one retained weighted contribution.
\label{ass:semantics}
\end{assumption}

\begin{proposition}[Trace-exact replay after physical redaction]
Under \cref{ass:det,ass:checkpoint,ass:semantics}, replay from a physically redacted store produces a model-and-optimizer state bit-identical to the separately executed trace-preserving oracle.
\label{prop:exact}
\end{proposition}

\begin{proof}
Both executions begin at the same checkpoint state. Consider plan record $p_j$. For a retained identifier, oracle and replay read the same frozen token and label arrays. For a forgotten identifier, both executions substitute the same fixed dummy arrays and weight zero before any forgotten-row lookup; therefore physical absence of the forgotten row does not change the numerical microbatch. The record also fixes the same microbatch seed, float32 learning-rate value, logical optimizer-step index, and accumulation-end flag. By \cref{ass:det}, the resulting forward pass, summed loss, backward pass, and any optimizer transition are bit-identical. Thus the model-and-optimizer state after plan record $j$ is identical whenever it was identical before record $j$. Induction over all replayed plan records gives equality of the terminal model and optimizer states.
\end{proof}

The proposition defines the equality invariant for the declared trace counterfactual. The empirical contribution is testing whether the implementation satisfies that invariant under physical replay-store redaction and large transformer state. Because oracle and replay share substantial code, their equality does not rule out a common-mode bug in that shared transition logic. It does verify the released invariant across separate executions, reconstructed provenance, different data-store materializations, and frozen state hashes. The proposition also does not equate the trace target with $\Repack(\Pplan,\F)$; \cref{sec:results160} uses repacking only as a non-equivalence witness.

\subsection{Release-time fail-closed verification}

Publication-facing runs execute behind artifact locks. The release runner resolves human-readable model references to 40-character immutable commits and hashes prepared token stores file by file and by directory. It also checks out external research repositories at pinned commits and refuses to proceed if the lock changes. Each run records the resolved configuration, environment snapshot, GPU information, plan hash, WAL and manifest hashes, model and optimizer hashes, exactness comparisons, runtime, workflow identity, and compact evidence artifact. This integrity model supports reproducibility and detects modification relative to trusted recorded hashes; it is not cryptographic non-repudiation against a malicious operator who controls both an artifact and its recorded hash. Production deployments could add authenticated logging or external attestation.

\section{Experimental Protocol}
\label{sec:experiments}

\subsection{Research questions}

The frozen campaign addresses four questions:
\begin{enumerate}[leftmargin=1.5em,itemsep=0.25em]
\item Can no-deletion replay reconstruct an original gradient-based LLM training run exactly, including optimizer state?
\item Can a deletion replay remain exact after the requested token rows are absent from the materialized replay token store?
\item Does exactness survive changes in deletion geometry and transfer to larger or different model families?
\item What descriptive behavioral measurements are obtained when a frozen exact state is passed through a standardized LLM-unlearning evaluator?
\end{enumerate}
The campaign does not answer whether the method is exact under multi-GPU training, whether it dominates approximate unlearning methods, or whether it satisfies a legal erasure obligation.

\subsection{Study A: Pythia 160M policy and deletion geometry}
\label{sec:study160}

We use Pythia 160M~\citep{biderman2023pythia} from the frozen \texttt{step143000} revision and a fixed-length WikiText-103-derived causal-LM token store~\citep{merity2016wikitext}. The run uses BF16 on one NVIDIA RTX A6000, microbatch size 4, gradient accumulation 4, one epoch, AdamW with learning rate $10^{-5}$, weight decay 0.01, 5\% warmup, cosine decay, and checkpoints every 250 logical optimizer steps. The execution plan contains 5,064 microbatches. We evaluate four forget-set geometries: early 0.1\%, middle 1\%, late 1\%, and random 5\%.

For each scenario we execute three deletion policies from the same eligible checkpoint. \texttt{slot\_mask} is the trace-preserving physical-redaction replay. \texttt{filter} removes forgotten rows from the forward microbatch, changing batch shape. \texttt{repacked} constructs a new densely packed retained-data plan. Only \texttt{slot\_mask} is expected to equal the trace oracle.

\subsection{Study B: Pythia 2.8B parameter-scale check}

The 2.8B endpoint uses the Pythia 2.8B \texttt{step143000} revision on one NVIDIA A100-SXM4-80GB. To keep the effective accumulation size aligned with the smaller study while fitting memory, microbatch size is 1 and gradient accumulation is 16. The run contains 20,256 WAL records and 1,266 logical optimizer updates, with the same $10^{-5}$ peak learning rate, 0.01 weight decay, 5\% warmup, cosine schedule, BF16 dtype, and eager attention. We test identity replay and the canonical random-5\% deletion only. Filtering and repacking are intentionally not rerun at this scale because their divergence is a policy question already isolated in Study A.

\subsection{Study C: Llama 3.2 1B and TOFU}

To test a second model family and a benchmark-facing data format, we use Llama 3.2 1B Instruct~\citep{meta2024llama32} and TOFU~\citep{maini2024tofu}. The frozen source model revision is \hashstr{9213176726f574b556790deb65791e0c5aa438b6}. The artifact lock independently pins the TOFU dataset revision and the tokenizer/reference revisions.

The preparation pipeline loads the full 4,000-example TOFU dataset and reproduces the pinned OpenUnlearning Llama 3.2 chat-template semantics. All tokens outside the final assistant response receive label $-100$, so only that response contributes to training loss. Preparation fails on examples longer than the fixed 512-token sequence length, preventing silent truncation. The \texttt{forget10} split identifies 400 dataset examples. Five epochs yield 20,000 sample presentations, represented by 2,500 microbatch plan/WAL records. We train the model on one RTX A6000 using BF16 and eager attention. The run uses microbatch size 8, gradient accumulation 4, 625 logical optimizer updates, AdamW peak learning rate $10^{-7}$, 20\% warmup, and cosine decay. The frozen training log shows the cumulative mean supervised-token loss decreasing from 2.693 after 50 updates to 2.132 at update 625. This confirms nontrivial optimization but does not establish how strongly the pre-fine-tuning base acquired the specific forget-set facts, because base-model TOFU metrics were not frozen.

After freezing the original state and identity replay, we materialize a new replay token store that omits the 400 requested examples. The redacted store contains 3,600 dataset examples and preserves the answer-only label arrays. We then compare trace-preserving replay with the separately executed trace oracle.

\subsection{Study D: secondary pinned OpenUnlearning diagnostic}

We reconstruct the Llama original and exact-deletion states from frozen release evidence and require their canonical model hashes to match before export to Hugging Face checkpoint format. Evaluation uses OpenUnlearning commit \hashstr{4ad738aaf60f6a4385f6e2506d01da99e76c31f3}~\citep{dorna2025openunlearning}, the TOFU \texttt{forget10}/\texttt{retain90}/\texttt{holdout10} configuration, eager attention, and the official \texttt{retain90} checkpoint at revision \hashstr{7114300c0049527a71833f5683965c358ad9dcbf}. The evaluation run performs zero training passes and zero optimizer updates.

OpenUnlearning's TOFU Forget Quality is the two-sample Kolmogorov--Smirnov (KS) test $p$-value comparing the model's forget-set Truth Ratio distribution with that of the retain reference. A $p$-value is not an effect size. We therefore also compute the KS statistic $D$ directly from the already frozen per-example logs; this is post-processing of released evidence, not a new model run. Model Utility is a harmonic aggregate of nine retain, real-author, and world-fact metrics in the pinned configuration. Because the official retain reference is unmatched and the campaign froze no base-to-target behavioral change, we do not use Study D to claim behavioral-unlearning efficacy.

A narrow interoperability patch casts BF16 probability tensors to float32 immediately before NumPy conversion at two upstream metric sites. Because every BF16 value is exactly representable in float32, the cast changes only representation compatibility; numerical metric values remain unchanged. The evaluation artifact records the patch and its affected locations.

\subsection{Exactness metrics and evidence policy}

For exact runs we record complete model SHA-256, optimizer SHA-256, unequal tensor count, unequal element count, maximum absolute parameter difference, and L2 parameter difference. We label a run \emph{exact} only when model state is bit-identical and optimizer hashes match. Numerical closeness is not promoted to exactness. We report behavioral metrics descriptively from the single frozen TOFU evaluation and make neither seed-level generalizations nor causal attributions to the deletion semantics.

\section{Results}
\label{sec:results}

\subsection{Exact replay across three endpoints}

\Cref{tab:endpoints} summarizes the central state-level result. Identity replay is exact at every endpoint. Trace-preserving deletion replay from a physically redacted store is also exact at every tested endpoint. The result spans a 17.1$\times$ increase in model elements from Pythia 160M to Pythia 2.8B and a separate Llama/TOFU training pipeline.

\begin{table}[t]
\centering
\caption{Frozen state-exactness endpoints. ``Exact'' means zero unequal model elements and matching optimizer state. Each deletion replay token store omits the requested dataset rows.}
\label{tab:endpoints}
\small
\begin{tabularx}{\textwidth}{l l r X X}
\toprule
Endpoint & GPU & Model elements & Identity replay & Deletion replay \\
\midrule
Pythia 160M & RTX A6000 & 162,322,944 & Exact & Exact for 4 geometries \\
Pythia 2.8B & A100 80GB & 2,775,208,960 & Exact & Exact, random 5\% \\
Llama 3.2 1B + TOFU & RTX A6000 & 1,498,482,688 & Exact & Exact, \texttt{forget10} \\
\bottomrule
\end{tabularx}
\end{table}

The equality includes optimizer state, which matters because a model with equal weights but different optimizer moments would not define an equal training state. At the 2.8B endpoint, identity replay is exact across 388 tensors and all 2,775,208,960 model elements. The random-5\% deletion removes 1,013 of 20,256 records, and the redacted replay exactly matches the oracle with zero unequal tensors, zero unequal elements, maximum absolute difference 0, L2 difference 0, and an equal optimizer hash.

On Llama 3.2 1B, identity replay is exact across 147 tensors and 1,498,482,688 model elements. The TOFU deletion removes 400 of 4,000 records; the replay store contains 3,600 retained records, preserves the label masks, and reports no forgotten identifiers present. The redacted replay again has zero unequal tensors/elements and an optimizer hash equal to the oracle.

\subsection{Pythia 160M isolates the counterfactual choice}
\label{sec:results160}

The smaller endpoint lets us compare replay policies directly. \Cref{tab:geometry} shows that slot-preserving replay is exact for every tested deletion geometry, while both physical filtering and dense repacking diverge from the trace oracle. We report the raw L2 differences only as witnesses of non-equality, never as quality scores. The table also exposes the locality dependence of suffix replay: a late request can use a substantially later checkpoint, whereas the random 5\% request begins at step zero.

\begin{table}[t]
\centering
\caption{Pythia 160M deletion-policy matrix. Checkpoint $k$ is the state immediately before logical step $k$. ``Suffix'' is $(1266-k)/1266$, the fraction of logical optimizer steps remaining from that checkpoint. L2 values are non-equivalence witnesses, not performance scores.}
\label{tab:geometry}
\small
\begin{tabular}{lrrrlll}
\toprule
Scenario & Forget & Checkpoint & Suffix & Slot & Filter & Repacked \\
\midrule
Early 0.1\% & 20 & 0 & 100.0\% & \textbf{Exact} & L2 3.1943 & L2 3.2366 \\
Middle 1\% & 203 & 250 & 80.3\% & \textbf{Exact} & L2 1.7244 & L2 2.6036 \\
Late 1\% & 203 & 750 & 40.8\% & \textbf{Exact} & L2 0.3264 & L2 1.9461 \\
Random 5\% & 1,013 & 0 & 100.0\% & \textbf{Exact} & L2 3.2358 & L2 3.1935 \\
\bottomrule
\end{tabular}
\end{table}

This table shows why ``train on $\D\setminus\F$'' is too coarse to name the paper's exact target. All three deletion programs use the same retained records. Only the slot-preserving program reconstructs the declared trace. Removing rows before the forward pass changes the numerical microbatch; repacking changes the plan itself. Neither outcome is a failure of those alternative counterfactuals. They are evidence that \emph{counterfactual specification is part of the unlearning problem}.

\subsection{Replay cost is locality-dependent}
\label{sec:cost}

\Cref{tab:runtime} reports frozen wall-clock times. These experiments do not establish deletion-time scalability: the Pythia 2.8B random-5\% replay takes 3,136.20 seconds versus 3,182.83 seconds for the original continuation, and the Llama/TOFU replay takes 1,715.13 seconds versus 1,722.50 seconds. Both restart from step zero. The method therefore demonstrates \emph{billion-parameter state exactness}, not sublinear or inexpensive deletion.

This behavior is structural for suffix replay. Consider a one-pass execution with $T$ distinct presentation positions and $m$ affected positions sampled uniformly without replacement. If $K_{\min}$ is the earliest affected position, the standard discrete order statistic gives
\begin{equation}
\mathbb{E}[K_{\min}]=\frac{T+1}{m+1}.
\end{equation}
With an ideal checkpoint immediately before that position, the expected fraction of the execution that must be replayed is
\begin{equation}
\mathbb{E}\!\left[\frac{T-K_{\min}+1}{T}\right]
=\frac{(T+1)m}{T(m+1)}\approx\frac{m}{m+1}.
\label{eq:replay_fraction}
\end{equation}
Periodic checkpoints can only move the restart earlier. Thus even under ideal placement, one uniformly random affected position implies roughly 50\% expected replay, 20 imply 95.2\%, 203 imply 99.5\%, and 1,013 imply 99.9\%. The one-epoch Pythia 2.8B random-5\% experiment has $T=20{,}256$ and $m=1{,}013$; empirically its earliest affected logical step is 2, while the latest eligible frozen checkpoint is step 0. Checkpoint replay therefore favors localized late requests. Large dispersed forget sets approach full replay, which is a structural limitation of the present architecture.

\begin{table}[t]
\centering
\caption{Frozen runtimes in seconds. Oracle and redacted-replay values are for the released canonical deletion scenario at each endpoint; Pythia 160M uses random 5\% here.}
\label{tab:runtime}
\small
\begin{tabular}{lrrrr}
\toprule
Endpoint & Original & Identity & Oracle & Redacted replay \\
\midrule
Pythia 160M & 369.95 & 319.52 & 310.40 & 310.74 \\
Pythia 2.8B & 3,182.83 & 3,192.49 & 2,978.87 & 3,136.20 \\
Llama 3.2 1B + TOFU & 1,722.50 & 1,720.07 & 1,716.39 & 1,715.13 \\
\bottomrule
\end{tabular}
\end{table}

\subsection{Control-provenance storage}

The compact WAL is not the whole systems-storage cost. \Cref{tab:storage} reports the measured binary-WAL-plus-manifest footprint and normalizes it by one frozen base checkpoint that contains model and optimizer state. The ratio is small, but checkpoint availability is essential to the method. The compact release deliberately excludes intermediate checkpoint files, so the frozen evidence cannot support a complete \emph{total unlearning-infrastructure storage} ratio. We therefore do not estimate one.

\begin{table}[t]
\centering
\caption{Measured control provenance versus one base model-and-optimizer checkpoint. These ratios exclude additional periodic checkpoints and are not total infrastructure overhead.}
\label{tab:storage}
\small
\begin{tabular}{lrrr}
\toprule
Endpoint & WAL + ID manifest & Base checkpoint & Ratio \\
\midrule
Pythia 160M & 1,296,298 B & 649,351,018 B & 0.200\% \\
Pythia 2.8B & 4,101,882 B & 11,100,993,590 B & 0.037\% \\
Llama 3.2 1B + TOFU & 958,890 B & 4,943,319,014 B & 0.019\% \\
\bottomrule
\end{tabular}
\end{table}

The 32-byte figure applies only to each fixed binary WAL record. Pythia 160M stores 5,064 such records (162,048 bytes) plus a 1,134,250-byte ordered-ID manifest; Pythia 2.8B stores 20,256 records (648,192 bytes) plus a 3,453,690-byte manifest; Llama/TOFU stores 2,500 records (80,000 bytes) plus an 878,890-byte manifest.

\subsection{OpenUnlearning is a secondary descriptive diagnostic}
\label{sec:behavior}

The frozen OpenUnlearning run is not used as evidence that trace-preserving deletion causes effective behavioral forgetting. Two controls required for that interpretation are absent: the official \texttt{retain90} checkpoint is not a matched local retained-data run, and no pre-fine-tuning base-model TOFU evaluation was frozen. We therefore move the complete aggregate table and distribution plot to \cref{app:behavior} and report only the descriptive facts needed to characterize the artifact.

For the 400 per-example forget-set Truth Ratio scores, the original and trace-deletion distributions are nearly coincident in this frozen run: their direct two-sample KS statistic is $D=0.015$. Against the independently trained official \texttt{retain90} reference, $D=0.2075$ for the original target and $D=0.2100$ for the trace-deletion state, with the corresponding upstream Forget Quality $p$-values $5.97\times10^{-8}$ and $3.91\times10^{-8}$. The $p$-values quantify evidence against equality to that reference distribution; they do not measure the magnitude of forgetting. Because the reference is unmatched, we treat these numbers as descriptive and make no causal comparison between trace preservation and retained-data retraining.

\section{Discussion}
\label{sec:discussion}

\subsection{The counterfactual is part of the specification}

Many definitions of unlearning write a reference learner as $A(\D\setminus\F)$, which is appropriate when $A$ denotes a complete randomized algorithm or a distribution over outputs. In an implemented deep-learning system, however, it is easy to slide between two different meanings of $A$: ``the same algorithm with the deleted rows removed'' and ``the same historical execution with those rows neutralized.'' Our results show that this distinction is not merely terminological. At Pythia 160M, dense repacking and physical filtering diverge from the trace oracle in every deletion geometry tested.

This does not make one counterfactual universally preferable. Dense retraining can be the appropriate reference when the policy goal is to approximate an independently retrained system. Trace preservation can be appropriate when the goal is to make a deployed, stateful training pipeline replayable and auditable under a specific removal semantics. The key requirement is to name the target before evaluating exactness.

\subsection{Optimizer state belongs in the exactness contract}

Machine-unlearning evaluations often report only model behavior or model parameters. Our release also hashes optimizer state. This is important for stateful training services: if the reconstructed model will continue training, identical weights with different Adam moments do not represent the same future learner. The distinction resembles the full-state versus observable-output criteria articulated by \citet{neel2021}. For our system, exactness is a claim about model-and-optimizer state under the recorded execution contract; the remaining transition controls are externalized into $\Pplan$ and $E$.

\subsection{Physical deletion and parameter evidence are different}

The trace-preserving plan retains logical identifiers, but the replay token store omits the corresponding token, label, and metadata rows. This operation establishes replay-store redaction only. It does not establish system-wide erasure from backups, caches, checkpoints, workflow artifacts, or other copies. The terminal parameter state likewise cannot prove that the data was never processed historically. Consistent with \citet{thudi2022}, the evidence bundle records an auditable procedure and its inputs. Parameter proximity alone cannot prove historical absence.

The manifest itself is provenance data and requires protection. In this research artifact it contains ordered sample identifiers for auditability. If an identifier encodes identity or sensitive metadata, retaining it may conflict with an operational deletion policy even after content rows are gone. Production systems can reduce that exposure with keyed pseudonymous markers and protected or signed manifests, potentially backed by external attestation.

\subsection{Behavioral diagnostics do not identify deletion semantics}

The OpenUnlearning run is useful as an interoperability and measurement record, not as an identified treatment effect. The direct original-versus-deletion Truth Ratio statistic is small in this frozen run, while both project states differ more from the independently trained official \texttt{retain90} reference. The frozen evidence lacks the matched retained-only run and base-to-full behavioral measurements needed to isolate the cause of that pattern. We therefore make no causal claim about trace preservation's effect on behavioral forgetting.

This restraint is consistent with recent work on benchmark limitations and suppression-versus-removal semantics~\citep{du2025false,xu2025reversibility,cooper2025position,yoon2026overused}. The behavioral appendix keeps the paper from conflating state exactness with a complete account of semantic forgetting.

\subsection{Where a trace-preserving contract can be useful}

The trace target serves systems that value reproducible post-deletion continuation under a declared historical execution contract. Examples include regulated stateful fine-tuning and deterministic release or audit workflows that must reconstruct a post-deletion state from an uncontaminated checkpoint without changing unrelated execution controls. In such settings, fixing order, RNG, schedule, and accumulation semantics removes one source of ambiguity from the deletion procedure.

The trade-off is prospective instrumentation and potentially large replay cost. A model trained without the required provenance cannot generally receive this guarantee retrospectively, and \cref{sec:cost} shows that dispersed requests tend to erase the computational advantage of checkpointing. The contribution is a precise systems primitive with a specific operating envelope. We make no claim that trace preservation dominates other exact-unlearning architectures or dense retained-data retraining.

\subsection{Relationship to exact-unlearning architectures}

SISA, S3T, LMEraser, and exact ICL unlearning reduce deletion cost by changing the learning architecture so influence is localized or adaptation is easily recomputed~\citep{bourtoule2021,chowdhury2025s3t,xu2025lmeraser,muresanu2025}. Trace-preserving replay makes a different trade-off. It stores compact control provenance but, for dispersed requests, can require essentially the entire continuation to be replayed. It preserves conventional gradient-based adaptation and avoids serving an ensemble of shards. That trade-off is worthwhile only when reconstructing a declared historical execution justifies the replay cost.

\section{Limitations and Claim Boundary}
\label{sec:limitations}

We scope the evidence to prospective, single-GPU reconstruction of a declared deletion counterfactual.

\paragraph{Single-GPU deterministic environments.}
All exactness claims are from pinned single-GPU runs. The Pythia 160M and Llama experiments use RTX A6000 GPUs; the Pythia 2.8B scale check uses an A100-SXM4-80GB. Distributed tensor, pipeline, or data parallelism introduces additional collective ordering and sharding state. We have not demonstrated multi-GPU exactness.

\paragraph{Environment dependence.}
We claim bitwise equality only for the recorded software, hardware, dtype, attention implementation, and deterministic-kernel settings. We do not claim equality across arbitrary PyTorch/CUDA versions or GPU architectures. The use of BF16 also means the claim is equality in the actual stored numerical representation, not equality to a real-arithmetic idealization.

\paragraph{One frozen seed per scientific endpoint.}
The state-equality tests are deterministic identity tests and do not require statistical inference to establish equality for a given run. However, broader conclusions about training dynamics, performance distributions, or behavioral metrics would benefit from multiple independent seeds. We do not make seed-level inferential claims.

\paragraph{Behavioral evidence is descriptive only.}
The official OpenUnlearning \texttt{retain90} checkpoint comes from separate training, and the frozen campaign contains no pre-fine-tuning base-model TOFU evaluation. Consequently, the behavioral measurements in \cref{app:behavior} do not establish how strongly the forget-set information was acquired or attribute any behavioral difference causally to trace preservation. We deliberately exclude behavioral efficacy from the paper's headline claims.

\paragraph{No publication-authoritative approximate baseline comparison.}
The repository contains integration hooks for established approximate methods. Because the release contains no frozen matched runs, we report no GradAscent, GradDiff, NPO, SimNPO, or other method rankings and make no superiority claim over approximate LLM-unlearning methods.

\paragraph{One-shot deletion only.}
Each frozen deletion experiment serves one request against the recorded source trace. Although $\F$ could syntactically be a union of identifiers, we have not tested an operational sequence $\F_1,\F_2,\ldots$ with checkpoint retirement, cumulative replay, provenance purging, or old replay-store cleanup. MUSE and sequential-deletion hooks in the repository are not evidence in this paper.

\paragraph{Prospective continuation and fine-tuning scope.}
The Pythia experiments start from frozen Pythia checkpoints and run controlled causal-LM training on a prepared WikiText trace corpus; the Llama experiment fine-tunes a frozen Llama 3.2 1B Instruct base on TOFU. We install the execution contract before these continuations begin. The results establish billion-parameter replay mechanics in these regimes, not retrospective exact deletion for an arbitrary pre-existing model and not end-to-end replay of a frontier model's original multi-node pretraining run.

\paragraph{Checkpoint storage and replay cost.}
Control provenance is small relative to one base checkpoint, but total infrastructure storage also includes the eligible checkpoint set. The compact release intentionally omits large intermediate checkpoints, so this paper does not report a complete storage-overhead ratio. Large dispersed deletion sets can also force almost full suffix replay, as quantified in \cref{sec:cost}.

\paragraph{Non-adversarial provenance integrity.}
CRC and SHA-256 checks support corruption detection and reproducibility relative to trusted hashes. They do not provide non-repudiation, an independently attested append-only audit log, or protection against an operator who can replace both an artifact and its hash.

\paragraph{Limits of replay-store redaction.}
The experiment proves absence of requested token, label, and metadata rows from the materialized replay token store. It does not prove deletion from backups, caches, original source stores, checkpoints, workflow artifacts, or every other storage layer. Trace exactness also does not imply that every semantic fact associated with a record becomes inaccessible or constitute a differential-privacy guarantee~\citep{shokri2017,carlini2021,du2025false}. We make no claim that the mechanism, by itself, satisfies GDPR Article 17 or any other legal obligation.

\section{Reproducibility and Artifact}
\label{sec:repro}

We release the code and frozen research records at \url{https://github.com/abdullah-x-bd/unlearning} as version \texttt{v0.3.1}~\citep{abdullah2026artifact}. The release commit is \hashstr{adf45d371909b14620b79350c8e748e9b3a995de}. The artifact lock records full model, dataset, and upstream repository revisions and prepared-data hashes. The publication-facing release directories record the compact evidence for Pythia 160M, Pythia 2.8B, Llama 3.2 1B + TOFU, and the downstream OpenUnlearning evaluation.

For the OpenUnlearning run, we pin the evaluator to commit \hashstr{4ad738aaf60f6a4385f6e2506d01da99e76c31f3}. The final result artifact records the original, exact-deletion, and reference evaluator logs, reconstruction summary, frozen hashes, GPU probe, allocation metadata, BF16 interoperability-patch provenance, and a manifest containing per-file sizes and SHA-256 values. The evaluation workflow reparses every JSON and verifies these hashes before upload. The evaluation metadata records zero training passes and zero optimizer updates.

Large model weights are not embedded in the compact GitHub evidence bundle. Instead, source model revisions and terminal model/optimizer hashes provide immutable identities. The release also includes a compact evidence bundle for durability beyond workflow-artifact retention windows.

\section{Conclusion}

Exact machine-unlearning claims need a counterfactual before they need an exactness metric. We define a deliberately narrow target: a prospectively instrumented trace-preserving continuation that assigns requested identifiers zero loss while replaying from an eligible uncontaminated checkpoint under the recorded execution controls and empty-segment optimizer semantics. Under pinned single-GPU environments, replay from token stores materialized without the requested rows is bit-identical to a separately executed trace oracle in model and optimizer state across Pythia 160M, Pythia 2.8B, and Llama 3.2 1B + TOFU.

The result is a systems construction with a narrow operating envelope. It establishes neither universal preference for this counterfactual nor cheap deletion. Randomly dispersed requests can force nearly full replay, and the method requires prospective provenance. We therefore keep standardized OpenUnlearning measurements as secondary descriptive diagnostics and draw no behavioral-efficacy conclusion from them. Replay-store redaction and state exactness do not by themselves establish behavioral forgetting. Connecting these properties requires matched experiments beyond the frozen campaign.

\bibliography{refs}
\bibliographystyle{tmlr}

\appendix

\section{Frozen Experimental Details}
\label{app:configs}

\subsection{Pythia 160M}

The frozen configuration uses seed 2026, strict deterministic algorithms, materialized redacted stores, BF16, eager attention, and the Pythia 160M \texttt{step143000} artifact resolved to commit \hashstr{b56d9bee36300031aeea723b73c4d62ac7fa71a2}. The fixed token store is \texttt{wikitext103-pythia-256}. Microbatch size is 4, gradient accumulation is 4, the run lasts one epoch, shuffle seed is 2026, and per-microbatch RNG derivation uses base seed 2027. AdamW uses peak learning rate $10^{-5}$, weight decay 0.01, 5\% warmup, and cosine scheduling. Checkpoints are written every 250 logical optimizer steps. The execution-plan SHA-256 is \hashstr{c617d628e7a04357f93e5994b8b68f67cd1cc14d3cdc8f7b815e99a258744a5e}.

Identity replay has zero unequal tensors and elements and matching optimizer state across 162,322,944 model elements. Every one of the four deletion scenarios has an exact slot-preserving replay and a physically redacted store with \texttt{forgotten\_ids\_present=false}. The compact evidence artifact is GitHub Actions artifact \texttt{9088115288}, SHA-256 \hashstr{f643f9baea2e819ac2494b605d11ee3dc15d991d9e1bd9bc5a9f6cd26eecba63}.

\subsection{Pythia 2.8B}

The scale configuration resolves Pythia 2.8B \texttt{step143000} to commit \hashstr{dbe7ae300a54abcdc475a33907b3dff81d25709f}. It uses BF16/eager attention, microbatch size 1, gradient accumulation 16, one epoch, peak learning rate $10^{-5}$, 0.01 weight decay, 5\% warmup, and cosine scheduling. The execution-plan SHA-256 is \hashstr{17ae171925ebd4b9c1beb393c7887d14b96ca09790da0b59a521fdf60198c431}.

The exact identity model SHA-256 is \hashstr{c68bcd5092e7c28dbb94e812f6ad62b64d4321bb4af0af3ef7784bb494048759}; the corresponding optimizer SHA-256 is \hashstr{fbaa0735422bc75e4855d5bb4e56b94d4cdc40a7a4c5a63b488f9a142379915f}. The random-5\% oracle/replay model SHA-256 is \hashstr{e23165a25eed20be896bd40ea3913dcf6e644c5f10ff54c16a35571a115df6f4}; optimizer SHA-256 is \hashstr{e7c6cce3619f0557275c8ae8e6a5db3b1e001ab868d844003c1ba17a1144ba89}. The evidence artifact is \texttt{9097605754}, SHA-256 \hashstr{50f7e82eed8d23ad64aa431d27f9dd91e400718f863860118f016b249de5d9c5}.

\subsection{Llama 3.2 1B + TOFU}

The source Llama revision is \hashstr{9213176726f574b556790deb65791e0c5aa438b6}. The TOFU dataset is pinned to \hashstr{324592d84ae4f482ac7249b9285c2ecdb53e3a68}. The prepared dataset directory SHA-256 is \hashstr{e1704f7f7073b892fe57ff423455f842b78e60ff03a4ee007505f6880f79360c}. The \texttt{forget10} identifier file SHA-256 is \hashstr{dc1f259e1c930087b81fc2c82ac38345a6ff1097760137bbb366a4f62230dc3d}.

The exact full-data target model SHA-256 is \hashstr{54c711e9bde77215d9c5def50429f925a382bdcd28150bb87a89a118dd54bc65}; optimizer SHA-256 is \hashstr{7dae55302988b834099f224a6ababcbd97a3411afa1b54e5e9a1739854f8e773}. The trace-preserving \texttt{forget10} oracle/replay model SHA-256 is \hashstr{067109bfd2e34f1616a8069d04ecd28b4814513332b03957ab917503122aeec3}; optimizer SHA-256 is \hashstr{0152484534eab198ebee5a500e77d81451f2cba34ea6baa987ccc5726e20daa1}. The exactness evidence artifact is \texttt{9104982056}, SHA-256 \hashstr{6a1381c76bec90233d5e4d1dde3fc43362405ed6bbd6d4378f709cdbf3634d8d}.

\section{Replay Algorithm}

\begin{algorithm}[H]
\caption{Trace-preserving physical-redaction replay}
\label{alg:replay}
\begin{algorithmic}[1]
\Require eligible checkpoint $\TState_k$ (state immediately before logical step $k$), verified plan $\Pplan$, redacted store $\D'$, forget-ID set $\F$, pinned environment $E$
\Ensure terminal state exact to the trace-preserving oracle under \cref{ass:det,ass:checkpoint,ass:semantics}
\State load $\TState\leftarrow\TState_k$
\State verify plan/WAL/manifest integrity and environment contract
\State initialize accumulated gradient state to empty
\For{each plan record $p_j=(I_j,s_j,\eta_j,u_j,a_j)$ at or after step $k$}
    \State reseed all configured RNGs with $s_j$
    \State set optimizer learning rate to $\eta_j$
    \State construct the microbatch in the original logical order
    \For{each identifier $i\in I_j$}
        \If{$i\in\F$}
            \State insert fixed dummy sequence with sample weight 0
        \Else
            \State read $i$ from the physically redacted store $\D'$
        \EndIf
    \EndFor
    \State compute sum-reduced weighted token loss and accumulate gradients
    \If{$a_j=1$}
        \If{the accumulation segment contains retained contribution}
            \State apply the optimizer update
        \Else
            \State skip the optimizer update \Comment{part of the declared counterfactual}
        \EndIf
        \State zero accumulated gradients
    \EndIf
\EndFor
\State hash terminal model and optimizer states and compare with oracle
\end{algorithmic}
\end{algorithm}

\section{Evidence Provenance}
\label{app:provenance}

\begin{table}[H]
\centering
\caption{Publication-facing workflow and artifact identities.}
\small
\begin{tabular}{llll}
\toprule
Evidence & Workflow run & Artifact ID & Artifact SHA-256 prefix \\
\midrule
Pythia 160M & 31451730675 & 9088115288 & \texttt{f643f9baea2e819a} \\
Pythia 2.8B & 31466226471 & 9097605754 & \texttt{50f7e82eed8d23ad} \\
Llama 1B + TOFU & 31490644488 & 9104982056 & \texttt{6a1381c76bec9023} \\
OpenUnlearning results & 31607929368 & 9147513923 & \texttt{b143e1784111ae3a} \\
OpenUnlearning metadata & 31607929368 & 9147515241 & \texttt{9f795ef861b03d53} \\
\bottomrule
\end{tabular}
\end{table}

The OpenUnlearning result artifact also contains an internal results TAR with SHA-256 \hashstr{d6b82c231fb73fecfa12996a4eb5bc68cdbb93959989ed660d7351464a6ce4ae}. The final evaluation ran on an NVIDIA A40 at a reported \$0.44/hour and reached cleanup after 2410.49 seconds, corresponding to an estimated GPU charge of \$0.2946. The metadata explicitly records zero training passes and zero optimizer updates. The behavioral results therefore evaluate the frozen canonical states; no checkpoint was newly optimized.

\section{Secondary OpenUnlearning Diagnostics}
\label{app:behavior}

This appendix reports the complete frozen OpenUnlearning aggregates and the effect-size information recoverable from the released per-example logs. We place these diagnostics outside the central exactness results because the official \texttt{retain90} reference is unmatched and the campaign froze no base-model TOFU evaluation.

\begin{table}[H]
\centering
\caption{Pinned OpenUnlearning TOFU aggregates. The official \texttt{retain90} model is an immutable calibration reference trained under a different trajectory, not a matched local retained-data control.}
\label{tab:ou}
\small
\begin{tabular}{lrrr}
\toprule
Metric & Original target & Exact deletion & Official \texttt{retain90} \\
\midrule
Model Utility & 0.354289 & 0.354136 & 0.592732 \\
Forget Quality ($p$) & $5.975\times10^{-8}$ & $3.913\times10^{-8}$ & -- \\
Forget Truth Ratio & 0.750241 & 0.750434 & 0.627597 \\
Forget QA probability & 0.168256 & 0.166909 & 0.116153 \\
Forget ROUGE & 0.344504 & 0.347532 & 0.379635 \\
Retain QA probability & 0.165048 & 0.164697 & 0.880278 \\
Retain ROUGE & 0.322241 & 0.322880 & 0.828251 \\
Retain Truth Ratio & 0.194750 & 0.195434 & 0.513091 \\
Min-K MIA AUC & 0.339891 & 0.336831 & 0.382331 \\
Extraction strength & 0.057374 & 0.056825 & 0.059611 \\
PrivLeak & 6.871098 & 7.366408 & 23.533750 \\
\bottomrule
\end{tabular}
\end{table}

\begin{figure}[H]
\centering
\includegraphics[width=0.68\textwidth]{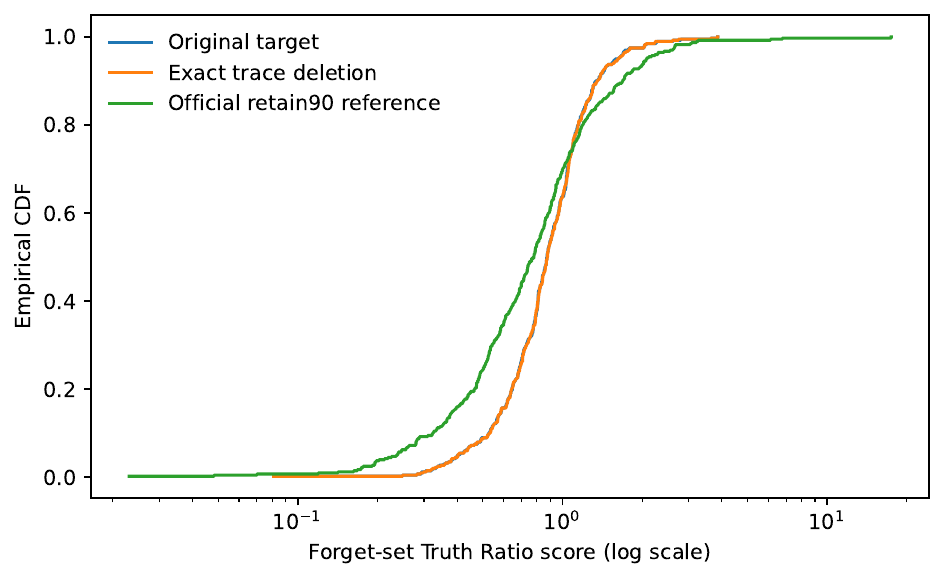}
\caption{Empirical CDFs of the 400 frozen per-example TOFU forget-set Truth Ratio scores. Original and exact-deletion distributions are visually close in this run; the official \texttt{retain90} reference is independently trained and is not a matched causal control.}
\label{fig:truthcdf}
\end{figure}

The upstream Forget Quality calculation compares 400 per-example forget-set Truth Ratio scores with the corresponding 400 scores from the official retain reference. Recomputing the KS statistic from those frozen scores gives $D=0.2075$ for the original target versus \texttt{retain90} and $D=0.2100$ for exact deletion versus \texttt{retain90}; these reproduce the released $p$-values above. A direct descriptive comparison of original and exact-deletion scores gives $D=0.015$. Their medians are 0.8729 and 0.8763, respectively, versus 0.7788 for the official reference. \Cref{fig:truthcdf} shows the empirical distributions. These statistics characterize the frozen artifact; they do not identify the causal effect of trace deletion.

The pinned OpenUnlearning implementation computes TOFU Forget Quality as a two-sample KS-test $p$-value. A small $p$-value provides evidence that sampled score distributions differ. Its magnitude does not measure how much a model has forgotten. Model Utility is the harmonic aggregate configured over retain, real-author, and world-fact submetrics. We report upstream aggregates as produced by the pinned evaluator.

The compatibility patch used for the release changes only two NumPy conversion sites in probability metrics: BF16 tensors are cast to float32 immediately before conversion. The patch leaves model weights, prompts, datasets, probability formulas, TOFU splits, and aggregation definitions unchanged. The result artifact records the exact source locations and patch.

\section{What Is Not in the Frozen Evidence}

For clarity, we exclude several repository implementations and experiment hooks from the paper's results because the frozen release does not contain their evidence. These include provenance-field ablations, checkpoint-spacing sweeps, duplicate-closure studies beyond the released deletion geometries, scaled XOR rollback, cohort-scoped LoRA experiments, curvature-based approximate deletion, MUSE, and publication-authoritative OpenUnlearning approximate-method baselines. Source-code presence alone provides no empirical evidence. We restrict the paper's quantitative claims to the frozen release records described above.

\end{document}